\documentclass{article}

\PassOptionsToPackage{authoryear, round}{natbib}

\usepackage[preprint]{nips_2018}

\usepackage[utf8]{inputenc} 
\usepackage[T1]{fontenc}    
\usepackage{hyperref}       
\usepackage{url}            
\usepackage{booktabs}       
\usepackage{amsfonts}       
\usepackage{nicefrac}       
\usepackage{microtype}      
\usepackage{graphicx}
\usepackage{subcaption}
\usepackage{setspace}
\usepackage{amssymb}
\usepackage{amsmath}
\usepackage{algorithm}
\usepackage{algpseudocode}
\usepackage{mathtools}
\usepackage{booktabs}
\usepackage{bbm}
\usepackage[font={small}]{caption}
\usepackage{enumitem}
\usepackage{cleveref}
\usepackage{wrapfig}

\usepackage[colorinlistoftodos]{todonotes}

\newcommand{\cutsectionup}{\vspace*{-0.08in}}
\newcommand{\cutsectiondown}{\vspace*{-0.05in}}

\newcommand{\cutsubsectionup}{\vspace*{-0.05in}}
\newcommand{\cutsubsectiondown}{\vspace*{-0.05in}}

\title{Many-Goals Reinforcement Learning}

\author{
  Vivek Veeriah \\
  University of Michigan \\
  Ann Arbor, MI 48109 \\
  \texttt{vveeriah@umich.edu} \\
  \And
  Junhyuk Oh \\
  University of Michigan\\
  Ann Arbor, MI 48109  \\
  \texttt{junhyuk@umich.edu} \\
  \And 
  Satinder Singh \\
  University of Michigan\\
  Ann Arbor, MI 48109  \\
  \texttt{baveja@umich.edu} \\
}

\begin{document}

\maketitle

\begin{abstract}
All-goals updating exploits the off-policy nature of Q-learning to update all possible goals an agent could have from each transition in the world, and was introduced into Reinforcement Learning (RL) by \citet{Kaelbling-allgoals}. In prior work this was mostly explored in small-state RL problems that allowed tabular representations and where all possible goals could be explicitly enumerated and learned separately. In this paper we empirically explore $3$ different extensions of the idea of updating many (instead of all) goals in the context of RL with deep neural networks (or DeepRL for short). First, in a direct adaptation of Kaelbling's approach we explore if many-goals updating can be used to achieve mastery in non-tabular visual-observation domains. Second, we explore whether many-goals updating can be used to pre-train a network to subsequently learn faster and better on a single main task of interest. Third, we explore whether many-goals updating can be used to provide auxiliary task updates in training a network to learn faster and better on a single main task of interest. We provide comparisons to baselines for each of the $3$ extensions.  
\end{abstract}

\cutsectionup
\section{Introduction}
\cutsectiondown

Off-Policy learning ~\citep{precup2001} is essential to  continual learning agents ~\citep{ring1994} for it allows simultaneous learning about many diverse goals using a single stream of experience generated by a behavior policy. This idea of ``all-goals updating''
was first demonstrated by \citet{Kaelbling-allgoals}, wherein a tabular Q-learning agent 
learned many goal-conditional action-value functions using a single stream of experience. Upon experiencing a transition from state $s$ to state $s'$ upon action $a$, the action-value functions are updated as follows:
\[
\forall g, \,\, Q_{g}(s,a) \leftarrow (1 - \alpha) Q_{g}(s,a) + \alpha[r_g(s,a) + \gamma \max_b Q_{g}(s',b)],
\]
where $Q_{g}(\cdot,\cdot)$ is the action-value function for goal $g$, $r_g(s,a)$ is the state-action reward function associated with goal $g$ and $\gamma$ is the discount factor. Because Q-learning is an off-policy algorithm the above many-goals update is sound in that with lookup table representations and appropriate conditions on the learning rate $\alpha$, the action-value functions for all goals would converge to their optimal values with probability one provided the behavior policy visited every state-action pair infinitely often in the limit of experience.



The above all-goals updating idea was demonstrated in small lookup table settings where all the goals were known in advance and a separate action-value function was learned for each goal. In this paper our main contributions are in the following \textit{three extensions} of Kaelbling's all-goals updating idea. 



\noindent{\bf 1) Unsupervised Mastery:} The first extension is to a visual RL domain without the use of lookuptable representations and without apriori knowledge of all possible goals available to the algorithm. Our main objective here is to explore whether the relatively straightforward adaptation of all-goals updating with separate lookuptables for the goals to our many-goals updating with a single neural network across goals will achieve mastery through generalization provided by the neural network without adding any special drive for mastery. Note that we refer to this setting as ``unsupervised mastery'' because there is no extrinsically provided reward function or main task. One challenge the lack of a main task raises is what should drive the behavior. 
We provide details of our approach on this, along with our empirical results below that evaluate the degree of mastery achieved.



\noindent{\bf 2) Unsupervised Pre-training:} The second extension is motivated by the more usual presence of a main task of interest and we explore whether using many-goals updating for unsupervised mastery as pre-training (cf. ~\citet{agrawal2016}) can subsequently improve the agent's ability to learn to achieve more rewards in its main task. Specifically, in the initial pre-training phase the agent behaves and learns to master many goals in its environment, as in the first extension described above. After this pre-training phase, the agent learns on-policy on the given main task using the pre-trained neural network as the initial network to fine tune during learning. Below we compare the performance of our many-goals unsupervised pre-training 
to pre-training based on reward-prediction as well as to instead providing the equivalent extra amount of training on the main task itself. 


\noindent{\bf 3) Many-Goals Auxiliary Tasks:} The third extension is to use many-goals updating as auxiliary control tasks to aid learning of the main task (cf. ~\citet{mirowski2016,vanseijen2017,teh2017}). Of course, since we cannot use all possible goals in realistic domains because they are too many 
of them, a challenge we address below is in how to select the many goals to use as auxiliary tasks. We also compare our method to the use of other state of the art auxiliary tasks. 

A summary of our main empirical findings is that on many Atari games, many-goals learning used for both pre-training and as auxiliary tasks significantly outperforms the baseline A2C~\citep{mnih2016} as well as other approaches for pre-training and auxiliary tasks added to A2C. 


\cutsectionup
\section{Related Work}
\cutsectiondown


Each of our three many-goals extensions listed above have related work, summarized briefly below.


\noindent{\bf Unsupervised Mastery.} 
Recently, \citet{sukhbaatar2018} introduced an approach that enables the agent to learn about its environment using a form of self-play. However, they mostly focused on exploration rather than mastery of the environment. 
Another related work~\citep{held2017} uses a generative adversarial network to generate goals that are at a good amount of difficulty to generate behavior for learning. A fundamental limitation of both the above approaches is that the agent learns only one goal at a time using an on-policy learning objective. This type of on-policy approach can be inefficient for mastery as we show in Section~\ref{sec:mastery}.

\noindent{\bf Supervised Mastery.} 
In supervised mastery, the entire set of goals (or tasks) that the agent needs to master are pre-defined and given in the form of pseudo-reward functions. The Unicorn architecture~\citep{mankowitz2018} and the Intentional Unintentional (IU) agent~\citep{cabi2017,riedmiller2018} were demonstrated to learn multiple tasks simultaneously using shared experience. Unlike this prior work on supervised mastery, our first extension focuses on unsupervised mastery where the set of goals is unknown as discussed above. In addition, our other two extensions investigate how unsupervised mastery can be useful for a main task by using mastery as unsupervised pre-training and as auxiliary control tasks. 

\noindent{\bf Unsupervised Pre-training for RL.} 
\citet{shelhamer2017} compared a number of supervised learning objectives, including reward prediction to pre-train an RL agent in terms of how quickly the agent learns to solve a main task after pre-training. \citet{munk2016} looked at the benefit obtained by initializing and freezing the parameters of an agent using parameters obtained by predicting observation and reward at next time step. 
\citet{levine2016} looked at initializing an agent with a network that classified images from ImageNet~\citep{deng2009}. However, these previous works do not seek to pre-train agents with unsupervised and temporally-extended control tasks, which is what we accomplish through our many-goals learning for pre-training approach.

\noindent{\bf Auxiliary Tasks.} The idea of using auxiliary tasks to improve an RL agent's state representation by jointly learning to maximize extrinsic and pseudo-rewards was first formalized by ~\citet{jaderberg2017}. They introduced pixel-control and feature-control tasks for learning to maximize change in pixel intensities and feature activations learned by the neural network respectively. However, it is unclear how much the learned knowledge for controlling pixels and features is directly helpful to the main task that the agent has to solve. Our empirical comparisons on Atari games in Section~\ref{sec:auxiliary} show that many-goals learning to reach observations that the agent has seen is more helpful than pixel-control for accelerating learning of the main task. 



\paragraph{Hindsight Experience Replay (HER).}
Some aspects of our extensions of all-goals updating are related to HER~\citep{andrychowicz2017} in that it also uses off-policy methods and UVFA ~\citep{schaul2015} to learn simultaneously about many goals from a shared stream of experience to accelerate the learning of a main task. There are a couple of important differences between HER and our paper. One difference comes from comparing HER to our unsupervised mastery work. HER assumes that 1) the main task is to reach a particular goal state (say $g^*$), and 2) the goal state ($g^*$) is given in advance to the agent. This supervised knowledge allows the HER agent to generate reasonable behavior by acting according to the given goal state. In contrast, our mastery method is completely \textit{unsupervised} in that the agent should not only learn about many goals but also pick the goal for generating behavior that makes mastery more efficient. A second difference comes from comparing HER and our work from the view of accelerating the learning of a main task (as opposed to mastery). We note that HER is not generally applicable to many RL domains, because HER requires a goal state $g^*$ to be specified and for example in Atari games there is no particular goal state (instead the objective is to maximize cumulative score).  In contrast, our method is not limited to such `goal-reaching' tasks, because we use many-goals learning for pre-training and separately as auxiliary tasks, both of which do not make any assumption about the form of the main task and are thus applicable to Atari games.

\begin{figure}
\begin{subfigure}{.5\textwidth}
  \centering
  \includegraphics[height=30mm]{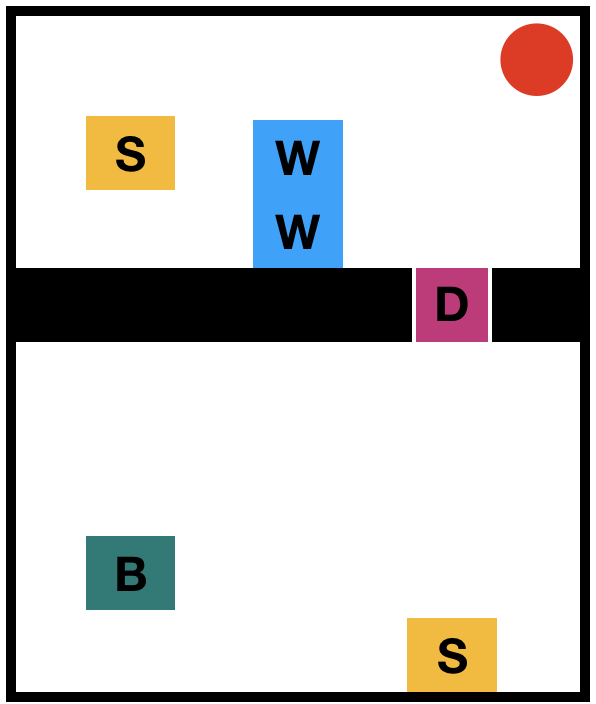}\label{fig:mastery_domain}
\end{subfigure}%
\begin{subfigure}{.5\textwidth}
  \centering
    \includegraphics[height=35mm]{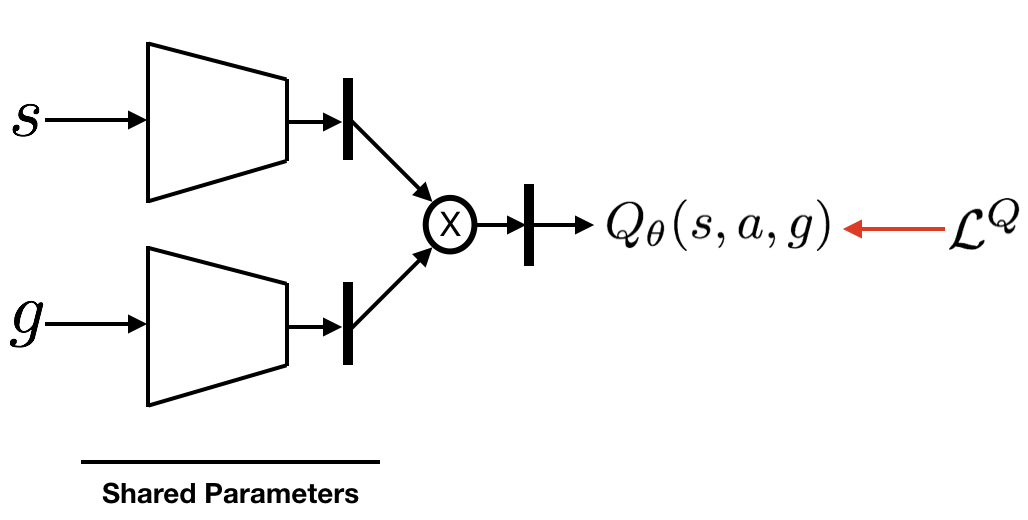}\label{fig:mastery_network}
\end{subfigure}%
\caption{{\em Left}: The visual gridworld domain used in our unsupervised mastery experiments. 
More details in text. {\em Right}: Neural network architecture for unsupervised mastery. The network parameters for producing the observation and goal embeddings are shared. The activations from these embeddings are used to produce $ Q_{\theta}(s, a, g) $ after combining them through multiplicative interactions.}
\label{fig:mastery_domain_network}
\vskip -0.2in
\end{figure}

\begin{algorithm}
\vskip -0.02in
\small
  \caption{Mastery of many goals}\label{alg:many_goals_mastery}
  \begin{algorithmic}
      \State Initialize parameter $ \theta $, replay buffer $ \mathcal{D} \gets \emptyset $
      \State Initialize goal buffer $ \mathcal{G} = \{s\}$ by collecting observations from a random policy
      \For{each iteration $i$}
      \State Sample a goal $g$ from $\mathcal{G}$ according to priority $p(g)$ (Eqn.~\ref{eqn:action_value_error})
     \State $ t \gets 0 $
     \While{the episode has not terminated}
      		\State $ s_{t} \leftarrow $ observe state
            \State $ a_{t} \sim \pi_{\theta}(a_{t} | s_{t}, g) $
            \State $ s_{t + 1} \leftarrow $ perform $ a_{t} $ in the environment
            \State Store transition $ \mathcal{D} \gets \mathcal{D} \cup \{ (s_{t}, a_{t}, s_{t + 1}) \} $
            \State Collect goal $\mathcal{G} \leftarrow \mathcal{G} \cup \{ s_{t + 1} \} $
            \State Sample a mini-batch of goals $\{g\}\in\mathcal{G}$ and transitions $\{(s,a,s')\} \in \mathcal{D}$
            \State Update the parameter $\theta \gets \theta - \alpha \nabla_\theta \mathcal{L}^{Q} $ (Eqn.~\ref{eqn:dqn_update})
            \State Update sampling priority $ p(g) $ for goals $g$ using Eqn.~\ref{eqn:action_value_error}
           \State $ t \gets t + 1 $
            \If{$s_{t + 1} == g$ or $ t == T $}
            	\State Terminate current episode
            \EndIf
      \EndWhile
      \EndFor
  \end{algorithmic}
\end{algorithm}

\cutsectionup
\section{Unsupervised Mastery in a Visual Gridworld} \label{sec:mastery}
\cutsectiondown

Here, our objective is to explore how generalization afforded by a neural-network representation of an all-goals (i.e., universal) action-value function aids unsupervised mastery, i.e., the learning of all possible goals without the presence of an extrinsic reward function or a main task. 

\noindent{\bf Domain.} We use the visual RL gridworld domain shown in ~\Cref{fig:mastery_domain_network} (Left). It consists of two rooms separated by a wall. Each room has a switch (S) that toggles the open/closed state of the door (D) that lets the agent (shown as a circle) move between rooms. The agent can move in any of the four cardinal directions; when it is at the switch location it has the additional toggle action available. The agent's actions are deterministic in all locations other than when it is at the dark blue square locations (W) where the agent's action becomes stochastic. Also, the door, once opened, stochastically switches back to being closed with probability $0.01$ at each time step. When the agent moves to the location of the block (B), the block moves in the direction of motion and thus the agent can push the block around. 
The agent's observations are the raw pixel-image of the gridworld (e.g, the image seen in ~\Cref{fig:mastery_domain_network} (Left)). The total number of possible visual observations is determined by combinatorics of the location of the agent, the location of the block (B), and the state of the door (D). 

\noindent{\bf Potential Goals.} 
For the purposes of this paper, we define a goal as a desired raw-pixel observation, and is said to be achieved when the current observation of the agent is identical to the goal observation. The $ 5728 $ feasible observations are the set of potential goal states. When training an action-value function for a goal image we will use a reward function for that goal, which is a small negative value for every transition to an observation that is not the goal observation and is zero for a transition to the goal observation. Since the set of goals $\mathcal{G}$ is initially unknown, the agent starts with an empty set of goals and adds to it as the agent discovers new observations by interacting with the environment.



\begin{figure}
\hspace{-0.1in}
\begin{subfigure}{.3\textwidth}
  \centering
  \includegraphics[width=49mm]{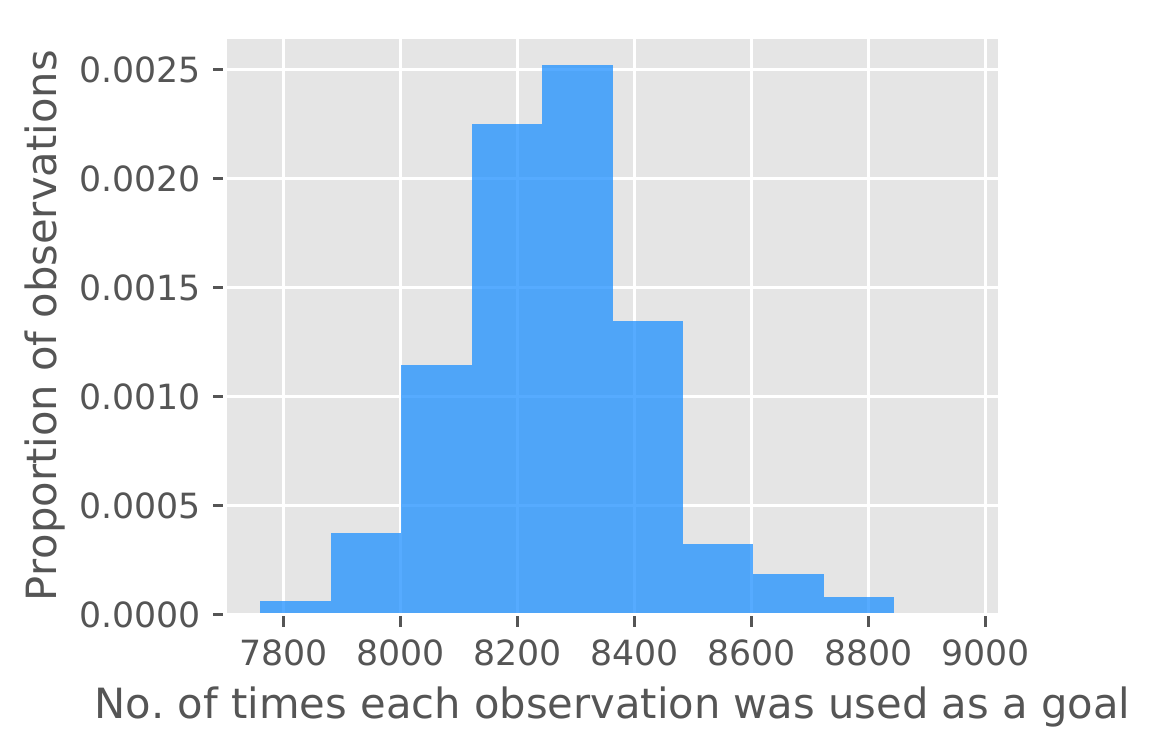}\label{fig:histogram_goals_updated}
\end{subfigure}%
\begin{subfigure}{.4\textwidth}
  \centering
  \includegraphics[width=44mm]{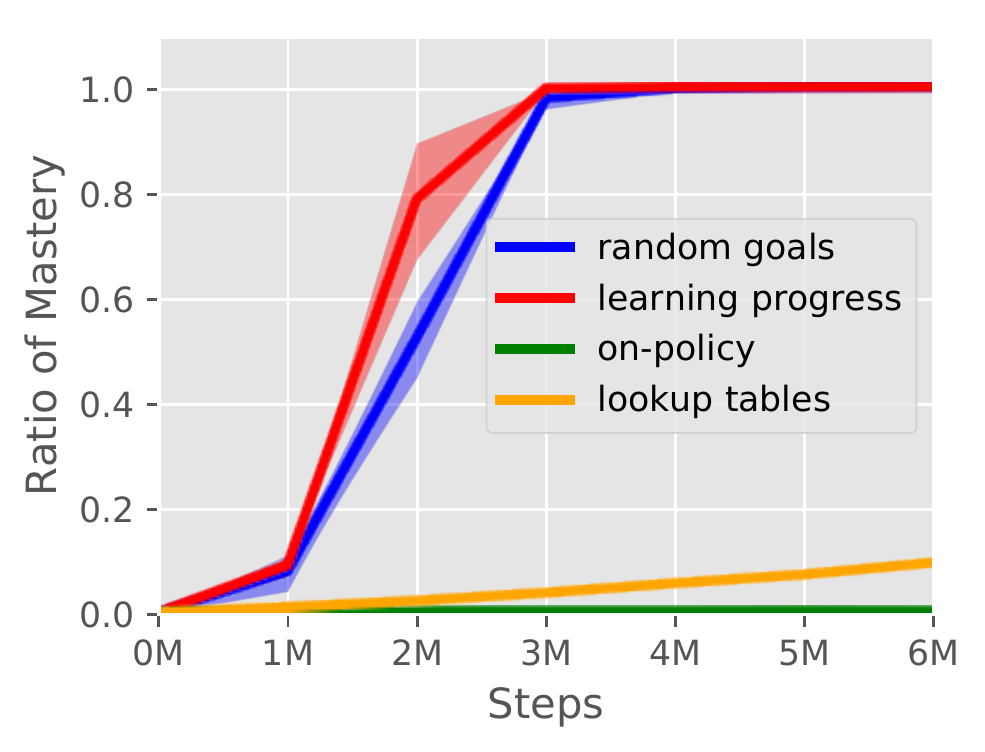}\label{fig:mastery_all_goals_plot}
\end{subfigure}%
\hspace{-0.2in}
\begin{subfigure}{.3\textwidth}
  \centering
  \includegraphics[width=49mm]{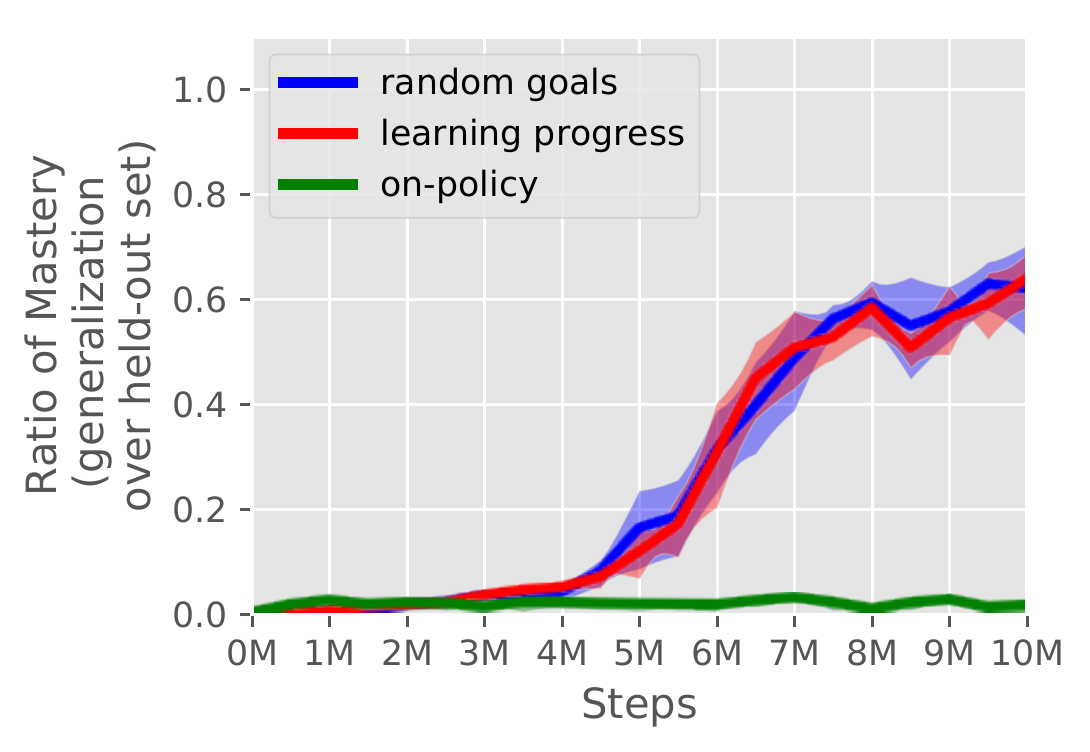}\label{fig:mastery_generalization_plot}
\end{subfigure}%
\caption{{\em Left}: Histogram showing the number of times each observation was used as a goal for updating the neural network. x-axis shows the bins and y-axis shows the proportion of observations falling into each bin. {\em Center}: Degree of mastery achieved over all possible goals with no held-out goals during training as a function of number of training steps. {\em Right}: Degree of mastery on held-out goal observations as a function of number of training steps. Center and Right figures show the mean performance obtained from $ 6 $ random seeds.} %
\label{fig:mastery_fig}
\vskip -0.1in
\end{figure}

\noindent{\bf Learning-Progress Based Behavior-Goal Prioritization.} Since our many-goal learning agent will learn simultaneously about many different goals from a single stream of experience, the behavior used for generating this experience plays a crucial role in the agent's efficiency in achieving mastery. We develop a method based on measuring \textit{learning-progress}, inspired from \citet{oudeyer2007,graves2017}, for prioritizing goals for selection at the start of an episode and compare it against choosing goals \textit{uniformly randomly}; both methods are constrained to only select previously seen observations as goals. 
More specifically, in the learning-progress method the priority of a goal $g$, denoted by $p(g)$, at an episode $i$ is defined as follows:
\begin{align}
p(g) &= \sum^{i}_{k=i-m} \mathcal{L}_{k-m}(g) - \mathcal{L}_{k}(g),  
\label{eqn:action_value_error}
\end{align}
where $\mathcal{L}$ is the squared Bellman error for a goal $ g $ defined in Equation~\ref{eqn:goal-loss} and $ m $ is a hyperparameter that controls how many of the statistics computed further back in time are used to compute this progress signal ($ m = 5 $ for our experiments). Intuitively, a positive priority means that the Bellman error has decreased over recent iterations. These priority values over goals are normalized to form a distribution that is sampled whenever a goal is chosen to generate behavior for an episode. 

\noindent{\bf Learning.}
We learn a goal-conditioned or universal action-value function, $Q_\theta(s,a,g)$, parameterized as a neural-network with weights $\theta$ as seen in~\Cref{fig:mastery_domain_network}(Right). The network consists of several shared layers that map observation $s_t$ and goal $g_t$ into the same embedding (i.e., feature) space. These two embeddings are combined together through multiplicative interactions to produce action-value estimates conditioned on a goal. 

As the agent interacts with the environment over an episode using a goal selected as described above (either randomly or using learning-progress), the state transitions $(s,a,s')$ are stored in the replay buffer $\mathcal{D}$. At the same time, unique observations are collected in the goal buffer $\mathcal{G}$. At every time step of interaction with the environment, $ 32 $ transitions are sampled from the replay buffer and 
$ 16 $ goals are sampled uniformly randomly from the goal buffer, in order to optimize the following objective:
\begin{align}
\mathcal{L}^Q &= \mathbb{E}_{g\sim \mathcal{G}} \left[ \mathcal{L}(g) \right]
\label{eqn:dqn_update}
\\
\mathcal{L}(g) &= \mathbb{E}_{s,a,s'\sim \mathcal{D}} \left[ r_g + \gamma_g \max_{a'}Q_{\theta^-}(s', a', g) - Q_\theta(s, a, g) \right]^{2}
\label{eqn:goal-loss}
\end{align}
where $r_g$ is a reward associated with $g$, which is $0$ when $s' = g$, and otherwise it is $ -0.1 $. Similarly, the discount factor $\gamma_g$ is $0$ when $s' = g$ and otherwise it is set to $ 0.99 $. $\theta, \theta^{-}$ correspond to the parameters of the learning network and the target network respectively. The overall method is summarized in Algorithm~\ref{alg:many_goals_mastery}.
\cutsubsectionup
\subsection{Experiment Design} 
\cutsubsectiondown

We conduct two different experiments. In the first, we let our algorithm choose goals from the observations seen thus far to drive behavior as well as choose many goals from the observations seen thus far for updating their action-value function. Here, we measure performance as the fraction of all possible goals learned by the agent as a function of the number of training steps. This experiment evaluates the ability of our behavior generation algorithm to drive the learning of mastery. In the second, we will apriori randomly partition the set of all goals into a held-out set of goals which the agent is prohibited from selecting for behavior generation as well as for many-goals updating. Here we measure performance as the fraction of held-out goals learned by the agent as a function of the number of training steps. This experiment evaluates the ability of our all-goals or universal action-value function to achieve mastery by generalizing to untrained goals.

\noindent{\bf Baselines.} One obvious baseline agent is an on-policy learning agent that uses exactly the same neural-network architecture as our agent, does exactly the same number of updates as our agent, chooses goals for episode generation uniformly random from the goal buffer, but restricts the updates to the same goal that was used to generate the latest episode. Another simple baseline agent is a tabular many-goals learning agent, which represents its goal-conditional action-values as separate lookup tables as in the original all-goals updating work by Kaelbling.  


\cutsubsectionup
\subsection{Results} 
\cutsubsectiondown
Figure~\ref{fig:mastery_fig} presents our results. We averaged results obtained from $ 6 $ random seeds for these plots. The figure on the left shows a distribution over how often goals are updated in the first experiment design; as expected this is roughly a bell-shaped curve, though perhaps unexpectedly the relatively tight spread over how often goals are updated means that all the observations were seen relatively early using the learning-progress method. The center figure compares mastery achieved by learning-progress versus random versus baselines (on-policy and tabular agent). Both the many-goals methods achieved complete mastery around $ 3 $ million steps, while the on-policy baseline struggles to achieve any level of mastery. Consistent with ~\citet{graves2017,mankowitz2018,sukhbaatar2018}, we find that random selection of goals to generate behavior is a hard baseline to beat though our learning-progress method does outperform it in terms of achieving higher mastery on average for training sizes smaller than $3$ million steps.
The figure on the right is from our second experiment, and measures the agent's generalization performance on a set of goals that were never used for training. Both the many-goals methods generalized to $ 60\% $ of goals that it had never used for training. In contrast to the results of Experiment $ 1 $, this $60\%$ mastery is entirely the result of generalization achieved by the universal action-value function. Consistent with our previous experiment, the on-policy baseline fails to achieve any generalization to the goals from the hold-out set.	

\begin{figure} \vspace{-1cm}
  \centering
    \includegraphics[height=42mm]{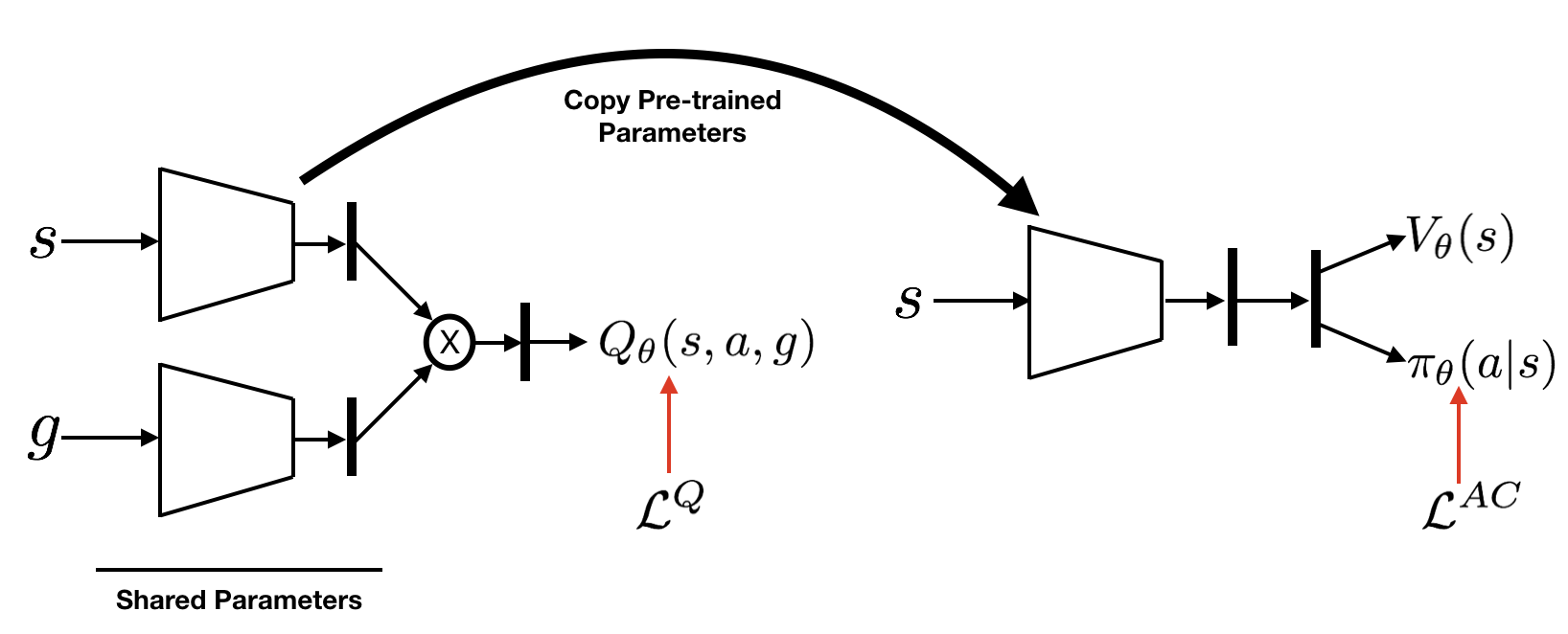}
\caption{Neural network architecture used for pre-training ({\em Left}), where the parameters are optimized using many-goals learning objective function. The parameters obtained after pre-training are subsequently used to initialize an agent ({\em Right}) that is fine-tuned through A2C learning updates.}
\label{fig:pre_trained_neural_net}
\vskip -0.1in
\end{figure}

\cutsectionup
\section{Mastery as Unsupervised Pre-training} 
\cutsectiondown
In this section, we evaluate our hypothesis that allowing the agent to learn to achieve many goals is a powerful pre-training approach for accelerating learning of the usual main task in $49$ Atari games~\citep{bellemare2013} (each with much larger observation spaces than we had in our visual gridworld above). Specifically, for each Atari game we do what we did in the Unsupervised Mastery section (i.e., Experiment $ 1 $ with random selection of goals) during the pre-training phase, and then in a fine-tuning phase evaluated if such pre-training helped accelerate the learning of the usual Atari game task relative to a baseline of using the same total amount of experience and training only on the main task as well as relative to another baseline of using a reward-prediction approach for pre-training.

\noindent{\bf Network Architecture.}
The neural network architecture used in our experiments for this section is shown in ~\Cref{fig:pre_trained_neural_net}.
The network architecture used for pre-training is identical to the one used for mastery. In the fine-tuning phase, the network learns a value ($V_\theta(s)$) and a policy ($\pi_\theta(a|s)$) as outputs, with its parameters initialized from those obtained from the pre-training phase (as visualized in the \Cref{fig:pre_trained_neural_net}). 

\noindent{\bf Learning.}
In the pre-training phase, the agent is trained through our mastery algorithm described in Section~\ref{sec:mastery} and Algorithm~\ref{alg:many_goals_mastery}. Goals in this setting are the observations (which are $4$ consecutive Atari frames). However, it is infeasible to prioritize goals based on the `learning progress' approach, 
because it requires computing the priority values for all goals collected in the goal buffer $\mathcal{G}$, which can be extremely large. Thus, we use `random goals' approach which samples goals uniformly from the goal buffer for behavior generation. In the fine-tuning phase, the agent is trained through advantage actor-critic (A2C) objective as follows:
\begin{align}
\mathcal{L}^{AC} &= \mathbb{E}_{s, a \sim \pi_{\theta}} \left[-\log\pi_{\theta}(a | s)(R-V_\theta(s)) + \beta \frac{1}{2} || R - V_{\theta}(s) ||^{2} \right]
\label{eqn:a2c_update}
\end{align}
where $R$ is an $ n $-step bootstrapped return and $ \beta $ is a hyperparameter that weights the influence of value function objective. We set $ n=5 $ and $ \beta=0.5 $ in our experiments.  

\cutsubsectionup
\subsection{Experiment \& Results}
\cutsubsectiondown
\begin{wraptable}{r}{6.5cm}
\centering
\small
\vskip -0.2in
 \begin{tabular}{c c c} 
 \toprule
 Agent & Median & >50\% \\
 \midrule 
 A2C (no-pre-training) & 32.8\% & 21 \\
 A2C + RP pre-training & 31.5\% & 18\\
 \midrule
 \textbf{A2C + MG pre-training} & \textbf{39.2\%} & \textbf{23} \\
 \bottomrule
\end{tabular}
\caption{Pre-training experiments: Summary of performance of agents on $ 49 $ Atari games trained for $ 10 $M steps ($ 1 $M steps of pre-training + $ 9 $M steps of fine-tuning) ($40$M frames). 'Median' shows median of human-normalized scores. '>50\%' shows the number of games where the human-normalized score is higher than 50\%.}
\label{tab:pre_training_results}
\vskip -0.2in
\end{wraptable}


\paragraph{Baselines.} A natural baseline for this scenario is an advantage actor-critic (A2C) agent that learns the given task from scratch without any form of pre-training. Another baseline for our experiments is an approach that pre-trains a learning agent to predict the sign of immediate reward from the environment given a sequence of previous transitions. This agent is called the reward-prediction agent and this form of pre-training was found to improve over an agent that learns from scratch \citep{shelhamer2017}.

\paragraph{Results.} 
The three agents namely, A2C agent, A2C agent with reward-prediction as pre-training and A2C agent with many-goals learning as pre-training are compared over $ 49 $ Atari games. The A2C agent was trained for $10$M training steps. For the pre-trained agents, the pre-training phase lasted $ 1 $M training steps. The resulting parameters were used to initialize the A2C agent, which was subsequently followed by $ 9 $M training steps involving A2C fine-tuning learning updates. For all our Atari experiments (both pre-training and auxiliary task) we averaged results obtained from $ 3 $ random seeds.

A short summary of the relative performance improvements achieved by different learning agents are shown in Table. \ref{tab:pre_training_results}. This table reports performance of agents by normalizing their scores with respect to the human player ~\citep{mnih2015}, the scores are called human-normalized scores. The game by game detailed relative performance improvements achieved by the agent that used many-goals learning for pre-training over its two baseline agents are shown in ~\Cref{fig:pre_train_result_1,fig:pre_train_result_2}. As can be seen in these figures, many-goals pre-training beat the baseline and the reward prediction in a majority of the games, in some cases by a significant amount.


\cutsectionup
\section{Many-Goals Learning as Auxiliary Tasks} \label{sec:auxiliary}
\cutsectiondown
In this section, we evaluate the use of many-goals learning as auxiliary tasks to accelerate learning of a main RL task.
The use of auxiliary tasks is an important research direction in DeepRL~\citep{jaderberg2017}. 
In this section, we introduce an approach that adaptively selects many goals for the agent to learn about, which are learned jointly with the main task.

\noindent{\bf Network Architecture.} 
The architecture for applying many-goals learning as auxiliary tasks to the learning agent is shown in ~\Cref{fig:aux_task_network}. The network consists of two branches: the main task branch consists of a value ($ V_{\theta}(s) $) and a policy ($ \pi_{\theta}(a|s) $) as outputs and the auxiliary task branch consists of Q-value estimates and separately receives a goal observation (sequence of $4$ frames) as input.

\noindent{\bf Learning.}
Similar to \citet{jaderberg2017}, the behavior of the agent is always produced by picking actions according to the main task policy $ \pi_{\theta}(a | s) $, while the auxiliary part of the network $ Q_{\theta}(s, a, g) $ is used only to improve the agent's representation of state. We maintain the $ K $-best episodes experienced by the learning agent in a replay buffer $ \mathcal{D} $. We sample trajectories of length $ n $ randomly from $ \mathcal{D} $ and update $Q_\theta(s,a,g)$ through many-goals learning, where the goal for each trajectory is set to be the last observation from the trajectory. Specifically, the objective is defined as a combination of actor-critic objective (Equation~\ref{eqn:a2c_update}) and off-policy many-goals learning objective (Equation~\ref{eqn:dqn_update}):$\mathcal{L} = \mathcal{L}^{AC} + \beta \mathcal{L}^{Q}$,
where $ \beta $ is a hyperparameter that weights the influence of the many-goals learning objective in the overall optimization. We tuned this hyperparameter on a fixed set of $ 7 $ Atari games, namely Atlantis, Boxing, DoubleDunk, Gravitar, Qbert, TimePilot, UpNDown and found $ \beta=0.02 $ to be optimal for our method. The results on the remaining games were generated using this fixed value. We set $ K=2000 $. 


\cutsubsectionup
\subsection{Experiment \& Results}
\cutsubsectiondown



\begin{figure}[ht]
\begin{minipage}[b]{0.4\linewidth}
\centering 
\small
\begin{tabular}{c c c} 
 \toprule
 Agent & Median & >50\% \\ 
 \midrule
 A2C & 32.8\% & 21 \\
 Pixel-control & 34.6\% & 20 \\
 Reward-prediction & 35.2\% & 21 \\ 
 \midrule
 \textbf{Many-goals} & \textbf{42.3\%} & \textbf{24} \\
 \bottomrule
\end{tabular}
\captionof{table}{Auxiliary task experiments: Summary of performance of agents on $ 49 $ Atari games trained for $ 10 $M steps ($40$M frames).}
\label{tab:aux_task_table}
\end{minipage}
\hfill
\begin{minipage}[b]{0.56\linewidth} 
\centering
\includegraphics[width=60mm]{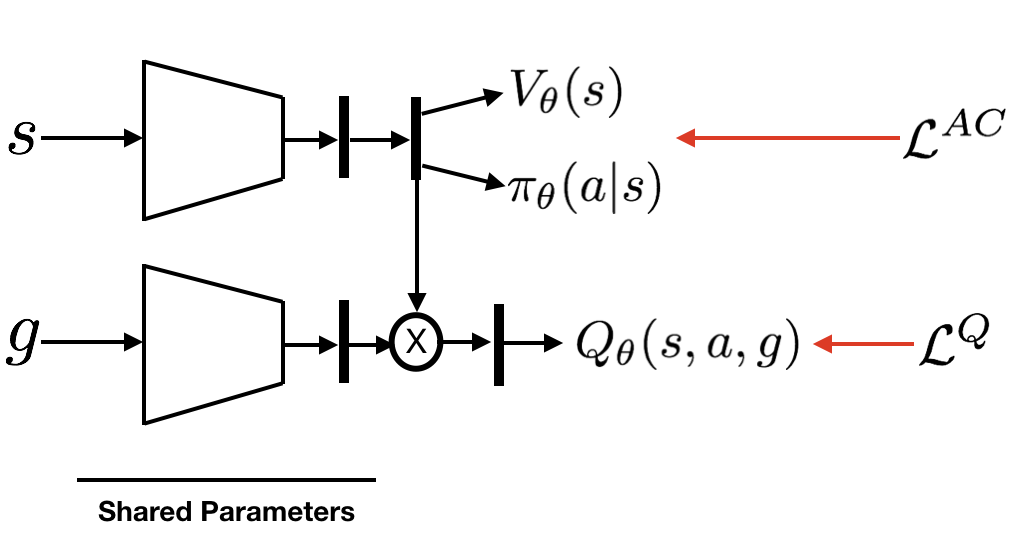}
\captionof{figure}{Neural network architecture for many-goals learning as auxiliary tasks.}
\label{fig:aux_task_network}
\end{minipage}
\end{figure}

\paragraph{Baselines.} 
We compare our approach 
against three different baselines: A2C, pixel-control agent ~\citep{jaderberg2017} and a reward-prediction agent ~\citep{jaderberg2017,shelhamer2017}.

\begin{figure}[p]
\begin{subfigure}[b]{\textwidth}
   \includegraphics[width=\textwidth]{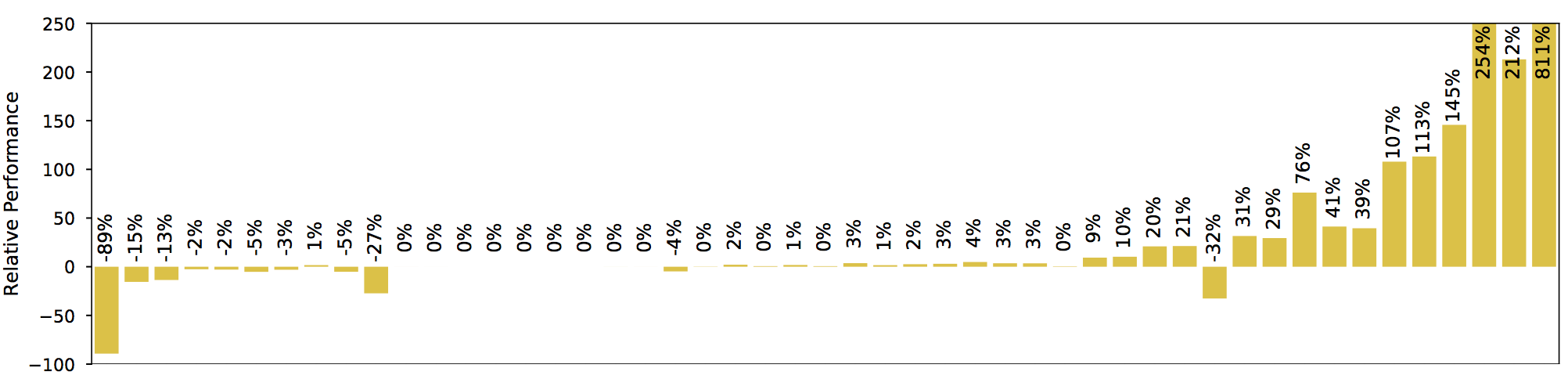}
   \vskip -0.05in
   \caption{Pre-training experiment: Many-goals as pre-training vs. no-pre-training A2C baseline}
   \label{fig:pre_train_result_1} 
\end{subfigure}

\begin{subfigure}[b]{\textwidth}
   \includegraphics[width=\textwidth]{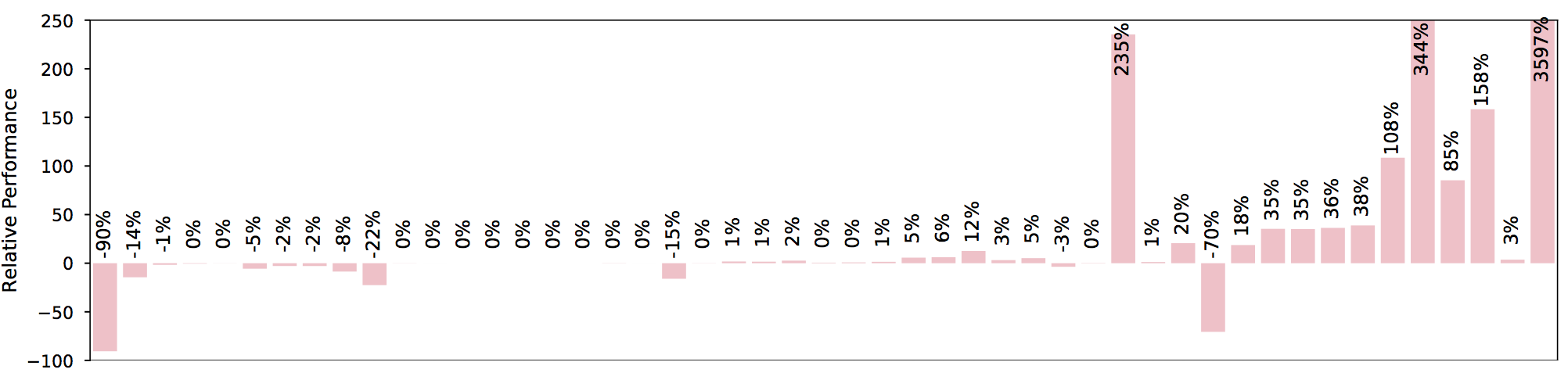}
   \vskip -0.05in
   \caption{Pre-training experiment: Many-goals as pre-training vs. Reward-prediction as pre-training baseline}
   \label{fig:pre_train_result_2}
\end{subfigure}

\begin{subfigure}[b]{\textwidth}
   \includegraphics[width=\textwidth]{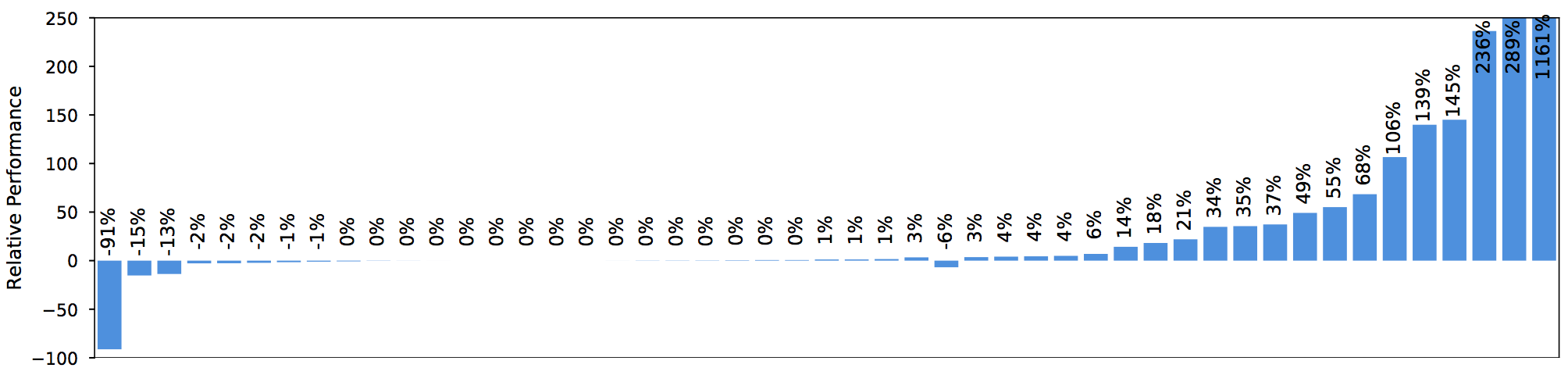}
   \vskip -0.05in
   \caption{Auxiliary task experiment: Many-goals vs. A2C baseline}
   \label{fig:aux_task_result_1}
\end{subfigure}

\begin{subfigure}[b]{\textwidth}
   \includegraphics[width=\textwidth]{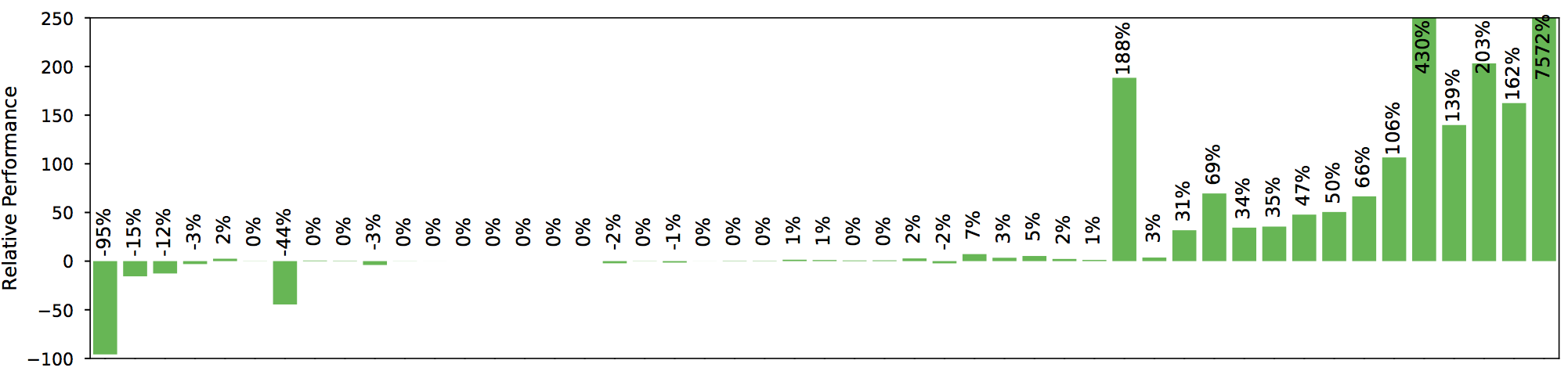}
   \vskip -0.05in
   \caption{Auxiliary task experiment: Many-goals vs. Pixel-control baseline}
   \label{fig:aux_task_result_2}
\end{subfigure}

\begin{subfigure}[b]{\textwidth}
   \includegraphics[width=\textwidth]{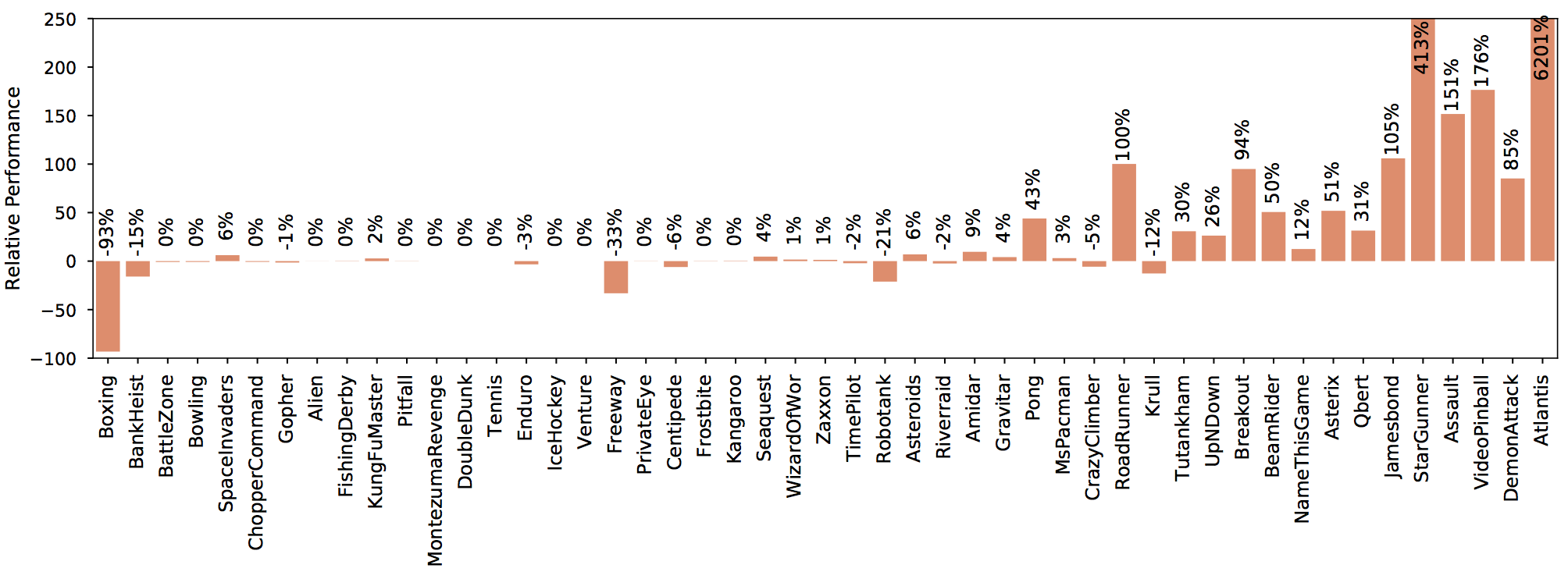}
   \vskip -0.05in
   \caption{Auxiliary task experiment: Many-goals vs. Reward-prediction baseline}
   \label{fig:aux_task_result_3}
\end{subfigure}
\vskip -0.05in
\caption{Relative performance improvements of many-goals over its baselines for pre-training and auxiliary task experiments on $ 49 $ Atari games. Plots show the mean performance obtained from $ 3 $ random seeds.}
\label{fig:relative_performance_plots}
\end{figure}

\paragraph{Results.}
We evaluate the performance achieved by different agents after $ 10 $M steps of training on $ 49 $ Atari games. The plots shown in~\Cref{fig:aux_task_result_1,fig:aux_task_result_2,fig:aux_task_result_3} show the relative performance improvements produced by many-goals learning as auxiliary task over the $3$ different baseline agents. \Cref{tab:aux_task_table} summarizes the human-normalized performance of different agents. From these, we see that for a majority of the games many-goals improves the agent's ability to accumulate more main-task rewards relative to all $3$ baselines (presumably due to better representation learning). 


\cutsectionup
\section{Conclusion}
\cutsectiondown
In this paper, we introduced three extensions to DeepRL of Kaelbling's all-goals updating algorithm. Our extension to unsupervised mastery showed that using a NN-based universal action-value function can lead to about $60\%$ mastery on held-out sets solely through generalization. Our extension of using many-goals updating for mastery as pre-training showed improved performance in a large majority of $49$ Atari games relative to both no pre-training as well as to using reward-prediction as pre-training. Our extension of using many-goals updating as auxiliary tasks showed improved performance in a large majority of $49$ Atari games relative to not using auxiliary tasks, to using pixel-control as auxiliary task, and to using reward-prediction as auxiliary tasks. Taken together, our results show great promise for the use of many-goals updating for a variety of important research goals for RL and Continual Learning including unsupervised mastery, pre-training, and auxiliary-tasks. 

\cutsectiondown

\clearpage
\section{Acknowledgments} This work was supported by a grant from Toyota Research Institute (TRI), by a grant from DARPA's L2M program, and by NSF Grant IIS 1319365. Any opinions, findings, conclusions, or recommendations expressed here are those of the authors and do not necessarily reflect the views of the sponsors.

\bibliography{nips_2018}
\bibliographystyle{plainnat}

\clearpage

\appendix
Here we present details for each of the three extensions presented in the main paper. 

\section{Unsupervised Mastery in Visual Gridworld}
Recall that we define goals as desired raw pixel observations. A goal is considered to be achieved when the current observation of the agent is identical to the goal observation.

\noindent{\bf Neural Network Architecture.} The network designed for the mastery agent allows the sharing of embeddings across both the observations and goals. The network consists of two convolutional layers, with $ 16 $ filters in the first layer with a stride length of $ 2 $ and $ 32 $ filters with a stride length of $ 2 $ in the second layer. The filter size for each of these convolutional layers is set as $ 2 \times 2 $. The fully-connected layer that immediately follows the convolutional layers consists of $ 512 $ units. The multiplicative interaction part of the network consists of embeddings of size $ 1024 $. All the activations were computed using Rectified Linear Units (ReLU). 
The inputs to the neural network consists of an RGB image (of size $ 10 \times 10 \times 3 $) that is produced as observation by the environment. 


\noindent{\bf Experiments.} 

We conducted two unsupervised mastery experiments: 1) where all goals were allowed during training, and 2) where some goals were held-out during training (We held-out a random set of 20\% of the $ 5732 $ goals for evaluation for each run). 

\noindent{\bf Baselines.} For our first experiment we compared against two baselines. 

\begin{itemize}[leftmargin=*]
\item {\bf On-policy baseline:} This agent learns {\em on-policy} about the goal that was used for generating experience. The architecture, learning algorithm, and the procedure for generating behavior remained the same as that of the many-goals learning agent. The only significant difference was in using the on-policy goal to make a learning update to the neural network. 
\item {\bf Tabular baseline:} The second baseline that we consider for this experiment is a tabular many-goals learning agent, which uses lookuptables to represent its goal-conditional action-value functions. More importantly, the state input for this tabular agent is the actual state of the environment, unlike the other agents which used the visual observation as input.
\end{itemize}

For our second experiment we only compare against the on-policy agent as the baseline because the lookuptable agent does not generalize by construction.  

\noindent{\bf Evaluation.} For measuring the mastery performance of different agents (for both our unsupervised mastery experiments), for each goal we evaluated whether the agent achieved that goal starting from a random start state within $ 200 $. If an agent failed to achieve a given goal within this time limit, that particular goal was marked as failure. We exhaustively evaluate from the set of all goals for the first experiment and from the set of held-out goals for the second experiment and report the average across this set. 

\noindent{\bf Hyperparameters.} The step-sizes for each approach was optimized by searching over the following range: $ \{ 1 \times 10^{-4}, 5 \times 10^{-4}, \cdots 0.01 , 0.05, 0.1\} $. RMSProp was used as the optimizer for updating the parameters of neural network. An $ \epsilon $-greedy policy over the action-value estimates is used for generating the agent's behavior. This exploration parameter $ \epsilon $ of the learning agent was annealed linearly from $ 1 $ to $ 0.1 $ over $ 1 $ million training steps, after which it remained at $ 0.1 $. The size of the replay buffer was set to $ 10,000 $. A batch of $ 32 $ transitions are sampled from the replay buffer and are modified according according a batch of $ 16 $ goals which are sampled independently from the goal buffer. In total, a batch of $ 32 \times 16 = 512 $ updates are made to the many-goal learning agent. For the on-policy baseline agent, a batch size of $ 16 $ transitions is used to make a learning update using the goal that is currently used for generating an episode.

\section{Mastery as Unsupervised Pre-training}

\noindent{\bf Neural Network Architecture.} The neural network architecture consists of three convolutional layers, each having $32, 64, 64$ filters respectively. The filter sizes for each layer were $ 8\times 8, 4\times 4, 3 \times 3 $ respectively and the stride lengths for each layer were $ 4, 2, 1 $ respectively. The convolutional layers were followed by a fully-connected layer with $ 512 $ units. The embeddings for both the observations and goals were shared. All the layers used ReLU as the activation function. 

During pre-training, the goal-conditioned action-value estimates were optimized using the many-goals learning objectives (refer main text). For the fine-tuning phase of the experiment, the parameters corresponding to the shared embeddings were used to initialize another neural network with same number of layers and specifications and was subsequently optimized using A2C objectives (refer main text). 

The inputs to all the agents consisted of $ 4 $ stacked Atari frames, where each Atari frame is a grayscale image of size $ 84 \times 84 $.  

\noindent{\bf Algorithm.} The algorithm for pre-training the agent through many-goals learning is exactly same as the one described to achieve unsupervised mastery in our main text.

\noindent{\bf Experiments.} For the pre-trained agents, the pre-training phase lasted $ 1 $ million training steps. During this phase, the learning agent performed many-goals training updates by selecting $ 16 $ transitions sampled randomly from the replay buffer and modified the reward for each transition using $ 5 $ goals sampled from the goal buffer. In total, during the pre-training phase, the many-goals agent performed $ 16 \times 5 = 80 $ learning updates. More importantly, the behavior for many-goals approach was generated by sampling random observations from the replay buffer and setting it as a goal for an episode. For the reward-prediction pre-training approach, experience was generated by picking uniformly random actions. At each training step, the reward-prediction agent sampled $ 80 $ transitions from the replay buffer and optimized its representation by classifying the sign of the reward experienced in each of the sampled transition. 

\noindent{\bf Baselines.} The baselines for evaluating our approach of using many-goals as pre-training is a conventional A2C agent that learns from scratch (i.e., without any form of pre-training) and a reward-prediction based pre-training approach ~\citep{shelhamer2017}. For the reward-prediction as pre-training approach, the agent's representation was pre-trained by predicting the sign of the immediate reward given a previous history of observations. As in our approach, the resulting pre-trained network after the reward-prediction task is used to initialize an A2C agent, which is then subsequently fine-tuned through on-policy A2C learning updates.

\noindent{\bf Evaluation.} We make a fair comparison of the three agents used in our pre-training experiments. Specifically, these three agents experience $ 10 $M steps from the environment ($ 40 $M Atari frames). For the pre-trained agents, the initial $ 1 $M training steps were performed using one of the previously described pre-training approaches, which is followed by $ 9 $M training steps where the agents are optimized using A2C learning updates. For the A2C baseline, there was no form of pre-training. This baseline agent performed $ 10 $M training steps directly optimizing its learning objectives.


\noindent{\bf Hyperparameters.} The $ \epsilon $ hyperparameter  that is used to generate an $ \epsilon $-greedy policy over the action-value estimated (during pre-training) was linearly annealed from $ 1.0 $ to $ 0.1 $ over the first $ 1 $ million steps, after which it is fixed to $ 0.1 $. The size of the replay buffer is set to $ 10,000 $. The step-size of the optimizer was set to $ 7e-4 $, which is the tuned step-size that was reported by the OpenAI baselines ~\citep{dhariwal2017} for Atari games. We used RMSProp as the optimizer for training the parameters of the neural network.

\begin{figure}[h]
\hspace*{-0.3in}
\centering
\includegraphics[width=\linewidth]{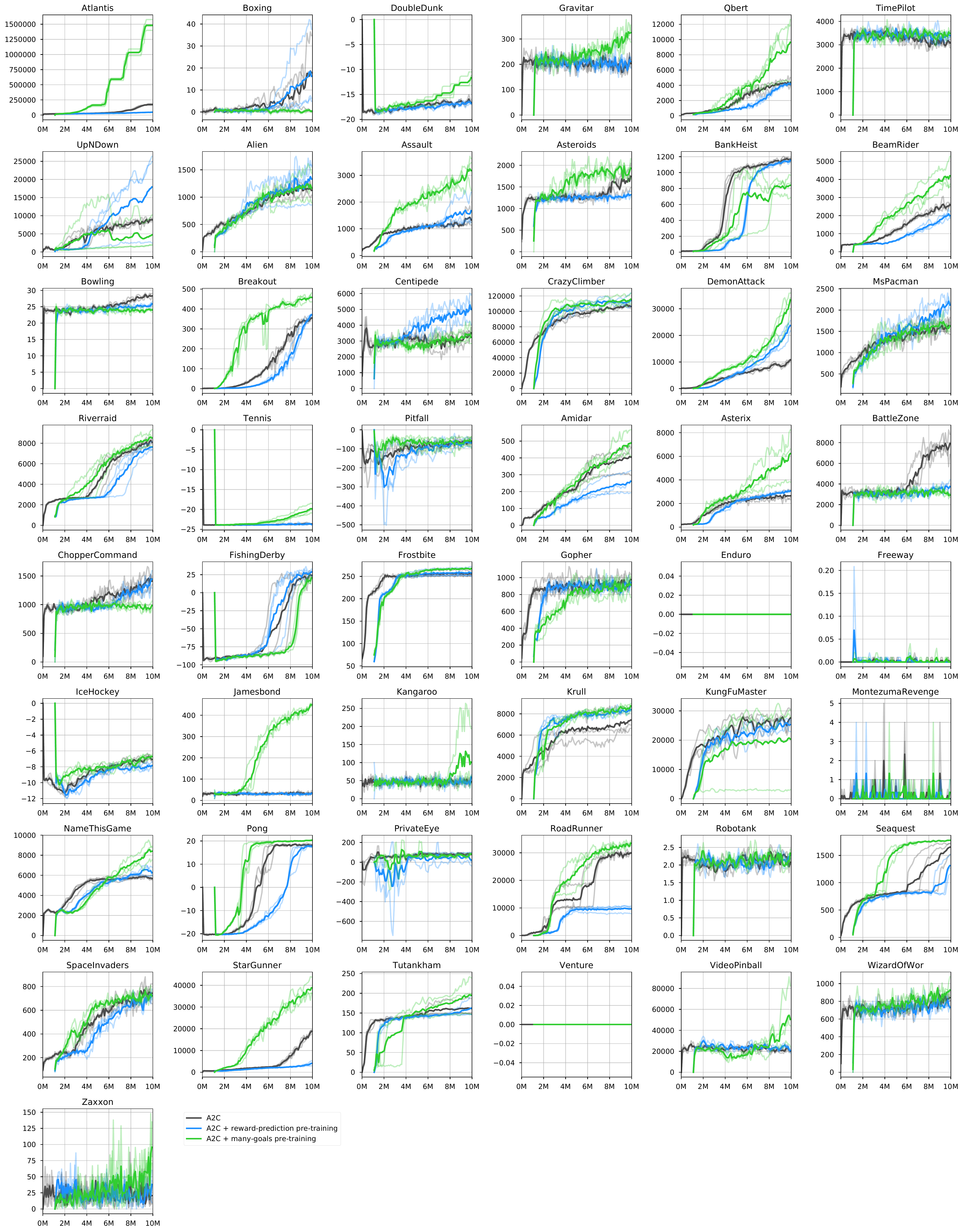}
\caption{Pre-training experiment: Learning performance of different agents on $ 49 $ Atari games}
\label{fig:atari_results_2}
\end{figure}

\noindent{\bf Results.} The learning curves for each Atari game from this experiment are shown in ~\Cref{fig:atari_results_2}. Also, the scores achieved by different agents in our pre-training experiments are reported in ~\Cref{tab:pretraining_scores}. All these plots and values are over $3$ runs using different random seeds. The results are obtained using $ 40 $M Atari frames. Note that the rewards achieved by the pre-trained agents were measured only during the fine-tuning phase. Specifically, the performance of the pre-trained agents were measured after a cumulative of $ 10 $ training steps (which includes $1$million pre-training steps.)




\begin{table}[p]
\centering
\begin{tabular}{ l| r r r }
\toprule
& A2C& A2C + RP & A2C + many-goals\\
\midrule
Alien& 1244.3& \textbf{1451.9}& 1354.4\\
Amidar& 413.9& 263.3& \textbf{494.9}\\
Assault& 1507.7& 1751.8& \textbf{3380.7}\\
Asterix& 2860.3& 3095.8& \textbf{6283.3}\\
Asteroids& 1800.6& 1391.6& \textbf{2123.6}\\
Atlantis& 173983.7& 47138.5& \textbf{1481710.7}\\
BankHeist& \textbf{1181.8}& 1151.7& 1000.2\\
BattleZone& \textbf{8736.7}& 4173.3& 3910.0\\
BeamRider& 2831.3& 2189.9& \textbf{4423.5}\\
Bowling& \textbf{29.0}& 26.1& 25.5\\
Boxing& \textbf{20.1}& 18.6& 2.2\\
Breakout& 359.7& 370.2& \textbf{472.7}\\
Centipede& 4306.7& \textbf{5404.8}& 3833.5\\
ChopperCommand& \textbf{1584.3}& 1463.8& 1103.3\\
CrazyClimber& 109049.2& 117350.5& \textbf{118226.0}\\
DemonAttack& 10766.4& 23947.6& \textbf{33374.4}\\
DoubleDunk& -16.2& -16.5& \textbf{-11.0}\\
Enduro& \textbf{0.0}& \textbf{0.0}& \textbf{0.0}\\
FishingDerby& 27.9& \textbf{31.6}& 21.7\\
Freeway& 0.0& \textbf{0.1}& 0.0\\
Frostbite& 258.8& 261.1& \textbf{268.9}\\
Gopher& \textbf{1067.1}& 1063.1& 1003.5\\
Gravitar& 263.4& 273.3& \textbf{352.5}\\
IceHockey& \textbf{0.0}& \textbf{0.0}& \textbf{0.0}\\
Jamesbond& 45.8& 43.4& \textbf{453.4}\\
Kangaroo& 72.7& 82.3& \textbf{134.7}\\
Krull& 7532.8& 8520.6& \textbf{8773.2}\\
KungFuMaster& \textbf{29788.0}& 27551.0& 21725.7\\
MontezumaRevenge& \textbf{3.3}& 2.0& 3.0\\
MsPacman& 1773.6& \textbf{2272.9}& 1837.2\\
NameThisGame& 5919.5& 6687.3& \textbf{8681.6}\\
Pitfall& -7.3& \textbf{0.0}& -8.2\\
Pong& 18.9& 17.9& \textbf{20.3}\\
PrivateEye& 120.9& 120.9& \textbf{133.9}\\
Qbert& 4520.4& 4292.7& \textbf{9771.8}\\
Riverraid& 8286.8& 7680.9& \textbf{8653.6}\\
RoadRunner& 30825.2& 10145.5& \textbf{33987.0}\\
Robotank& 2.5& 2.5& \textbf{2.7}\\
Seaquest& 1654.8& 1312.5& \textbf{1777.3}\\
SpaceInvaders& \textbf{824.9}& 768.0& 780.6\\
StarGunner& 18729.5& 4090.3& \textbf{39182.3}\\
Tennis& \textbf{0.0}& \textbf{0.0}& \textbf{0.0}\\
TimePilot& 3831.7& 3899.2& \textbf{3918.2}\\
Tutankham& 164.0& 163.0& \textbf{197.1}\\
UpNDown& 9575.6& \textbf{18049.6}& 6625.7\\
Venture& \textbf{0.0}& \textbf{0.0}& \textbf{0.0}\\
VideoPinball& 27836.6& 32157.8& \textbf{57335.2}\\
WizardOfWor& 931.5& 896.3& \textbf{1006.7}\\
Zaxxon& 70.7& 73.3& \textbf{124.7}\\
\bottomrule
\end{tabular}
\caption{Pre-training experiment: Scores achieved by different agents on $ 49 $ Atari games.}
\label{tab:pretraining_scores}
\end{table}

\section{Many-Goals Learning as Auxiliary Tasks}

\noindent{\bf Neural Network Architecture.} The neural network architecture for this experiment has the same specifications for convolutional and fully-connected as described in our pre-training experiment. A policy head and value head are connected over the main task branch of the network. All the activations were computed using Rectified Linear Units (ReLU). 

\noindent{\bf Algorithm.} The algorithm for using many-goals learning as auxiliary task is described in ~\Cref{alg:many_goals_aux_tasks}.

\begin{algorithm}[h!]
  \caption{Many-goals learning as auxiliary tasks}\label{alg:many_goals_aux_tasks}
  \begin{algorithmic}[1]
      \State Initialize parameter $ \theta $, replay buffer $ \mathcal{D} \gets \emptyset $
      \For{each iteration}
      		\State Initialize an episode buffer $ \mathcal{E} \leftarrow \emptyset $
                \For{each step}
                    \State $ s_{t} \leftarrow $ current observation from the environment
                    \State $ a_{t} \sim \pi_{\theta}(a_{t} | s_{t}) $
                    \State $ s_{t + 1}, r_{t + 1} \leftarrow $ perform $ a_{t} $ in the environment
                    \State Store transition $ \mathcal{E} \gets \mathcal{E} \cup \{ (s_{t}, a_{t}, r_{t + 1}) \} $
                \EndFor
			\If{$ s_{t + 1} $ is terminal}
                \State $ \mathcal{D} \leftarrow \mathcal{D} \cup \{ (s_{t}, a_{t}, r_{t + 1}) \} $ for all $ t $ in $ \mathcal{E} $
                \State Clear episode buffer $ \mathcal{E} \leftarrow \emptyset $
                \State Keep only $ K $-best episodes in $ \mathcal{D} $ 
            \EndIf
            \State \textit{\# Update $ V_{\theta}(s) $ and $ \pi_{\theta}(a | s) $}
            \State $ \theta \gets \theta - \alpha \nabla_{\theta}\mathcal{L}^{AC} $ 
            \State \textit{\# Update $ Q_{\theta}(s, a, g) $}
            \State $ \theta \gets \theta - \alpha \beta \nabla_{\theta}\mathcal{L}^{Q} $ 
      \EndFor
  \end{algorithmic}
\end{algorithm}

\noindent{\bf Experiments.} 
We jointly train the goal-conditioned action-value estimates (using many-goals learning objective); and the policy and value functions (using A2C objectives) at each training step. We measure the rewards achieved by the agent while following the policy that is being learned by the agent. All our Atari experiments were for $ 40 $M Atari frames. We used a batch size of $80$ for updating the auxiliary branch of the network.

\noindent{\bf Baselines.} The baselines for evaluating our approach of using many-goals as auxiliary task is a conventional A2C agent (without any auxiliary task), a pixel-control agent~\citep{jaderberg2017} and a reward-prediction agent~\citep{jaderberg2017,shelhamer2017}. 


\noindent{\bf Hyperparameters.} The hyperparameters specific to our method (which is $ \beta $) was tuned over this range: $ \{ 1 \times 10^{-4}, 5 \times 10^{-4}, \cdots 0.01, 0.05, 0.1 \} $ and the best value for this hyperparameter was found by evaluating our method on a fixed set of Atari games, namely Atlantis, Boxing, DoubleDunk, Qbert, Gravitar, TimePilot and UpNDown. The remaining hyperparameters, like the coefficients for the entropy regularizer, and value function loss were tuned for the A2C baseline agent and the same hyperparameters were used for our method as well as the baseline agents.  The step-size of the optimizer was set to $ 7e-4 $, which is the tuned step-size that was reported by the OpenAI baselines ~\citep{dhariwal2017} for Atari games. It is important to note here that we did not tune any hyperparameter other than $ \beta $ for all the agents that used auxiliary tasks, and more importantly, the previously described set of 7 games were used for tuning this hyperparameter. The hyperparameter $ K $ is set to $ 2000 $. RMSProp is used as the optimizer for optimizing the parameters of the neural network.

\noindent{\bf Evaluation.} We measure the reward achieved by the different agents on $ 49 $ Atari games over the number of training steps.

\begin{figure}[h!]
\hspace*{-0.25in}
\centering
\includegraphics[scale=0.3]{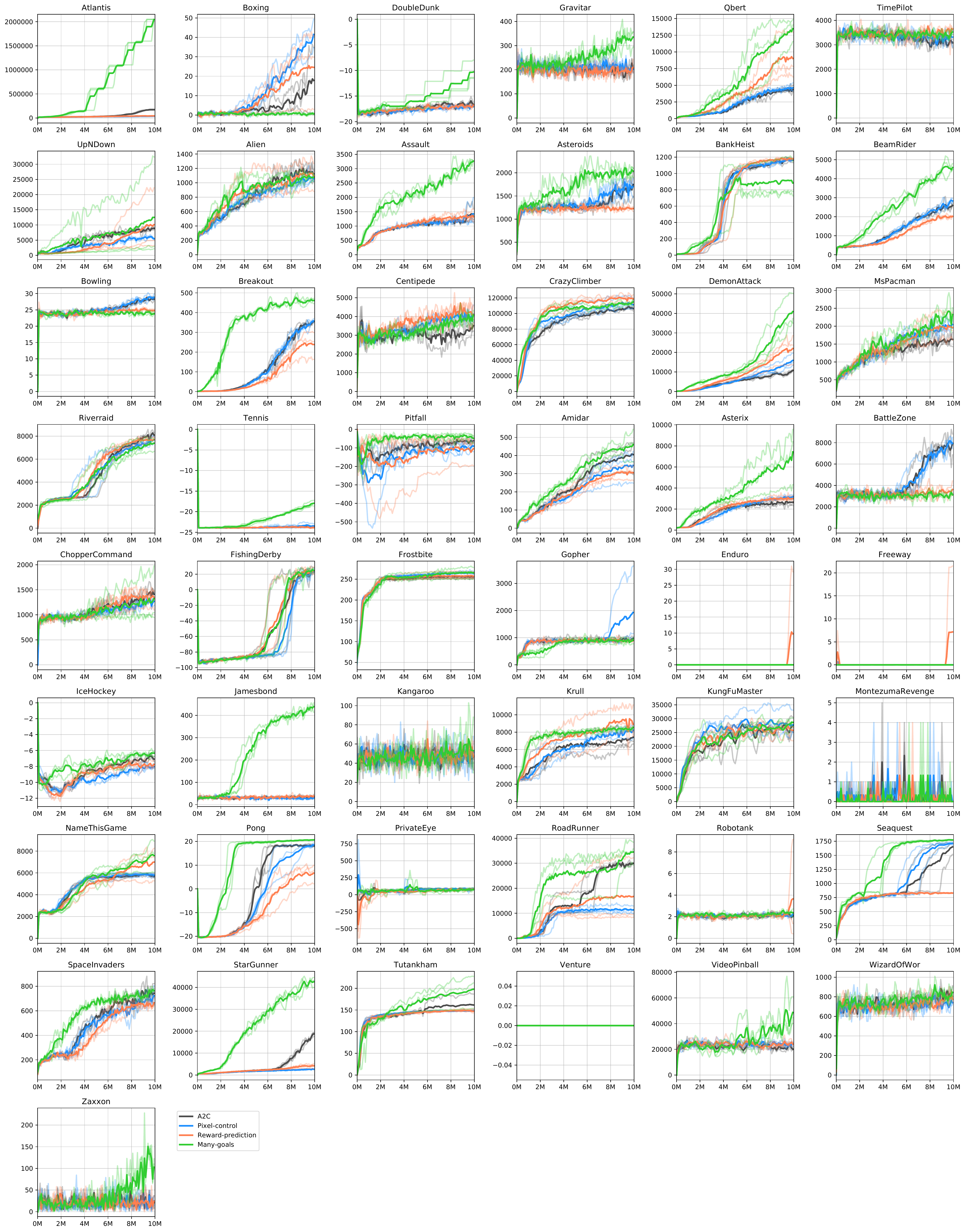}
\label{fig:aux_tasks_atari_performance}
\caption{Auxiliary task experiment: Learning performance on $ 49 $ Atari games}
\label{fig:atari_results_1}
\end{figure}

\noindent{\bf Results.}
The scores achieved by the agents in different games are tabulated in ~\Cref{tab:aux_tasks_score}. The learning curves for each method across Atari games are shown in ~\Cref{fig:atari_results_1}.





\begin{table}[p]
\centering
\begin{tabular}{ l| r r r r }
\toprule
& A2C& Pixel-control& Reward-prediction& Many-goals\\
\midrule
Alien& \textbf{1244.3}& 1136.1& 1180.6& 1175.9\\
Amidar& 413.9& 354.0& 314.1& \textbf{474.4}\\
Assault& 1507.7& 1535.7& 1440.8& \textbf{3372.1}\\
Asterix& 2860.3& 3243.2& 3129.2& \textbf{7429.7}\\
Asteroids& 1800.6& 1875.3& 1359.0& \textbf{2221.1}\\
Atlantis& 173983.7& 39340.5& 45105.3& \textbf{2045339.7}\\
BankHeist& 1181.8& 1186.7& \textbf{1191.0}& 1003.6\\
BattleZone& \textbf{8736.7}& 8368.3& 4188.3& 3875.0\\
BeamRider& 2831.3& 2928.4& 2110.3& \textbf{4848.7}\\
Bowling& 29.0& \textbf{29.3}& 26.4& 25.3\\
Boxing& 20.1& \textbf{42.1}& 25.9& 1.8\\
Breakout& 359.7& 362.6& 250.5& \textbf{486.8}\\
Centipede& 4306.7& 4475.7& \textbf{4929.5}& 4328.1\\
ChopperCommand& \textbf{1584.3}& 1366.5& 1451.0& 1379.8\\
CrazyClimber& 109049.2& 114590.7& \textbf{122306.8}& 115868.0\\
DemonAttack& 10766.4& 15892.0& 22453.0& \textbf{41450.7}\\
DoubleDunk& -16.2& -17.4& -17.1& \textbf{-10.5}\\
Enduro& 0.0& 0.0& \textbf{10.3}& 0.0\\
FishingDerby& \textbf{27.9}& 26.6& 26.8& 27.0\\
Freeway& 0.0& 0.7& \textbf{9.8}& 0.0\\
Frostbite& 258.8& 267.8& 260.4& \textbf{269.9}\\
Gopher& 1067.1& \textbf{1953.0}& 1064.7& 1034.4\\
Gravitar& 263.4& 280.0& 263.2& \textbf{366.5}\\
IceHockey& \textbf{0.0}& \textbf{0.0}& \textbf{0.0}& \textbf{0.0}\\
Jamesbond& 45.8& 46.1& 48.4& \textbf{448.5}\\
Kangaroo& 72.7& 75.0& 74.0& \textbf{85.0}\\
Krull& 7532.8& 8363.6& \textbf{9634.5}& 8612.3\\
KungFuMaster& 29788.0& \textbf{30948.3}& 28935.5& 29729.2\\
MontezumaRevenge& 3.3& \textbf{4.0}& 3.0& 3.0\\
MsPacman& 1773.6& 2187.0& 2058.2& \textbf{2532.7}\\
NameThisGame& 5919.5& 5952.9& 7103.2& \textbf{7702.7}\\
Pitfall& -7.3& -4.8& \textbf{-1.2}& -12.1\\
Pong& 18.9& 18.6& 7.5& \textbf{20.7}\\
PrivateEye& 120.9& \textbf{382.1}& 135.2& 247.4\\
Qbert& 4520.4& 4766.0& 9430.5& \textbf{13612.3}\\
Riverraid& \textbf{8286.8}& 7733.2& 7754.5& 7456.0\\
RoadRunner& 30825.2& 12219.3& 17600.3& \textbf{35216.3}\\
Robotank& 2.5& 2.6& \textbf{4.7}& 2.7\\
Seaquest& 1654.8& 1715.5& 844.2& \textbf{1777.5}\\
SpaceInvaders& \textbf{824.9}& 747.4& 692.7& 784.9\\
StarGunner& 18729.5& 2739.7& 4369.5& \textbf{44002.3}\\
Tennis& \textbf{0.0}& \textbf{0.0}& \textbf{0.0}& \textbf{0.0}\\
TimePilot& 3831.7& 3844.3& \textbf{3911.8}& 3860.7\\
Tutankham& 164.0& 148.7& 150.2& \textbf{198.3}\\
UpNDown& 9575.6& 6777.0& 10184.2& \textbf{12721.0}\\
Venture& \textbf{0.0}& \textbf{0.0}& \textbf{0.0}& \textbf{0.0}\\
VideoPinball& 27836.6& 29104.4& 30345.2& \textbf{55207.9}\\
WizardOfWor& 931.5& 895.8& 889.5& \textbf{957.3}\\
Zaxxon& 70.7& 79.2& 60.0& \textbf{180.0}\\
\bottomrule
\end{tabular}
\caption{Auxiliary task experiment: Scores achieved by different agents on $49$ Atari games.}
\label{tab:aux_tasks_score}
\end{table}



\end{document}